# On the Computability of Artificial General Intelligence

Georgios Mappouras
george.mappouras@gmail.com

Charalambos Rossides
rossides.ac@gmail.com

**Abstract**

*In recent years we observed rapid and significant advancements in artificial intelligence (A.I.). So much so that many wonder how close humanity is to developing an A.I. model that can achieve human level of intelligence, also known as artificial general intelligence (A.G.I.). In this work we look at this question and we attempt to define the upper bounds, not just of A.I., but rather of any machine-computable process (a.k.a. an algorithm). To answer this question however, one must first precisely define A.G.I. We borrow prior work's definition of A.G.I. [1] that best describes the sentiment of the term, as used by the leading developers of A.I. That is, the ability to be creative and innovate in some field of study in a way that unlocks new and previously unknown functional capabilities in that field. Based on this definition we draw new bounds on the limits of computation. We formally prove that no algorithm can demonstrate new functional capabilities that were not already present in the initial algorithm itself. Therefore, no algorithm (and thus no A.I. model) can be truly creative in any field of study, whether that is science, engineering, art, sports, etc. In contrast, A.I. models can demonstrate existing functional capabilities, as well as combinations and permutations of existing functional capabilities. We conclude this work by discussing the implications of this proof both as it regards to the future of A.I. development, as well as to what it means for the origins of human intelligence.*

## 1. Introduction

In recent years we have seen big leaps in the advancement of artificial intelligence (A.I.) technologies like autonomous vehicles, medical diagnosis, and others [2, 3, 4, 5, 6, 7, 8, 9]. More impressive though have been the advancements in generative A.I. Generative A.I. leverages large language models (LLMs) that can process queries described in natural language [10, 11]. Thus, they enable humans to interact with machines simply by expressing requests in natural language rather than software code. LLMs have demonstrated the ability to analyze such queries and generate accurate and coherent responses, with applications across various industry sectors [12, 13, 14, 15]. The accuracy of LLMs has improved dramatically showing an ability to "analyze" the input query and "reason" about the answer they should provide.

These improvements have led to the resurface of the age-old question *"Can machines achieve a human level of intelligence?".* Or more formally, can machines achieve Artificial General Intelligence (A.G.I.)? A.G.I. is an artificially (machine-generated) level of intelligence that matches the human level of intelligence, also known as General Intelligence (G.I.). Although prior work has proposed tests that evaluate the intelligence of machines and compare it with the human intelligence [1, 16, 17, 18, 19, 20], to the best of our knowledge, no definite, formal answer exists whether machines can achieve A.G.I. or not.

In this work we attempt to answer this exact question by providing a formal proof and giving a definite answer on this question. To do that, we must first define A.G.I.

### 1.1 Defining A.G.I.

The term A.G.I. is often used to describe a level of machine intelligence that enables cognitive functionalities similar to humans [21]. Such functionalities include reasoning, learning, creativity, etc. However, many A.I. companies give somewhat different definitions. We discuss these definitions to find common ground and conclude on a single A.G.I. definition that attempts to encapsulate the core, common principles.

Researchers at DeepMind propose a definition that requires an A.G.I system to meet six criteria [22]. Based on these criteria they define different levels of classification for A.G.I. Level 2, called "*Competent A.G.I.*" is the next level to be achieved where A.G.I. *"...must have performance at least at the 50th percentile for skilled adult humans on most cognitive tasks".* That means that A.G.I. should be able to perform better at reasoning and creativity (fundamental human cognitive abilities) than most humans.

OpenAI defines A.G.I. as *"...a highly autonomous system that outperforms humans at most economically valuable work"* [23]. Many studies conclude that innovation is the primary source of driving economic expansion [24, 25, 26]. Thus, an A.G.I. system based on OpenAI's definition should be able to surpass humans in driving innovation and thus, surpass human reasoning and creativity. This is further supported by Sam Altman's (CEO of OpenAI) comments in a recent interview with journalist Mathias Döpfner and physicist David Deutsch [27] where Sam Altman proposed that if a future A.I. model could *"...figure out quantum gravity and could tell you it's story"* that would be enough to prove that it has achieved A.G.I. In other words, an A.G.I. system should be able to showcase the ability to create new knowledge.

Although MetaAI and xAI do not provide a specific definition for A.G.I. in their websites, Mark Zuckeberg (CEO of Meta) and Elon Musk (founder of xAI) have publicly commented on their vision of A.G.I. [28, 29]. Their vision aligns with the definitions of A.G.I. we have already discussed where a machine can perform any intellectual task a human can, including tasks it was not specifically trained for. Other industry giants like IBM, Amazon's AWS, and Salesforce all give similar definitions for A.G.I. [30, 31, 32].

The common denominator across all these definitions is that a system that achieves A.G.I. should truly be creative and able to innovate. Therefore, to be able to define A.G.I. we must be able to define creativity. Sam Altman gave one such definition proposing that an A.I. would unequivocally be creative if it was able to *"...come up with new scientific knowledge"* [27]. That means that A.I. would be able to generate some new knowledge that was not known before and thus could not have been "coded in" the model through training.

Recent work on testing systems for A.G.I. [1] has generalized this idea by defining A.G.I. as the ability to create *new functionality* across any domain (scientific, arts, sports, etc.) that was not directly coded in the system through hard-coded function or training data. The *new functionality*, allows the system to output *new knowledge*. More specifically, an A.G.I. system analyzes some input (reasoning) and creates some *new knowledge* (creativity) that allows it to perform some *new functionality* (innovation) that was not previously possible. (e.g., as Sam Altman proposed, the ability to explain a new scientific theory not previously understood by humans).

To further explain this concept let us consider a schematic of a hypothetical A.G.I. system as seen in **Figure 1**. Initially the system has a set of functionalities $\mathbb{F}$. However, no function in that group can generate the output $\vec{Y}$ for an input $\vec{X}$. As we see in **Figure 1**, an A.G.I. system has a mechanism (depicted as "System's A.G.I. mechanism") that can analyze input $\vec{X}$ (and possibly other data from $\mathbb{F}$) and creates some *new functionality* $f_{n+1}(\vec{X}) \notin \mathbb{F}$. The new functionality is then added to $\mathbb{F}$ and can now be used to output the new result $\vec{Y}$. We use this ability as the definition of an A.G.I. system, and we formalize it in **Definition 1**.

As mentioned in [1], humans are indeed G.I. systems as they meet the requirements described in **Definition 1**. Humans continuously demonstrate the ability to generate *knew knowledge* that leads to *new functionalities* that were previously not possible. This is done through continuous technological and scientific advancements, as well as through the evolution of arts, sports, and culture. Thus an A.G.I. system, that has cognitive abilities comparable to humans, should also be able to meet the requirements of **Definition 1**.

## 1.2 Motivation

Having precisely defined A.G.I. in Section 1.1 we now motivate this work and explain the importance of answering the question if an algorithm can ever achieve A.G.I. We do so by investigating the implications of both hypothetical scenarios. That is, the scenario where some algorithm exists that can achieve A.G.I., as well as the scenario where no algorithm exists that can achieve A.G.I.

An algorithm that would achieve A.G.I., based on our definition in Section 1.1, would be able to create new functionalities and thus, be creative and innovative. Thus, a company or other entity that would yield such an algorithm would be able to significantly benefit economically, as the A.G.I. algorithm would drive economic growth through innovation. Furthermore, as machines can process information orders of magnitude faster than humans [33], the rate of innovation would likely be proportionally greater, leaving any competitors that do not utilize A.G.I. far behind in terms of economic and developmental growth. Thus, in this scenario the incentive to heavily invest in the research and development of an A.G.I. algorithm would be enormous. This

> **Definition 1.** A system is A.G.I. if and only if for some input $\vec{X}$ it can produce some output $\vec{Y}$ where the function that maps $\vec{X}$ to $\vec{Y}$, $f(\vec{X}) = \vec{Y}$ did not initially exist in the system's group of available functionalities $\mathbb{F}$. Rather the new functionality $f(\vec{X}) = \vec{Y}$ was generated through the A.G.I. mechanism of the system.

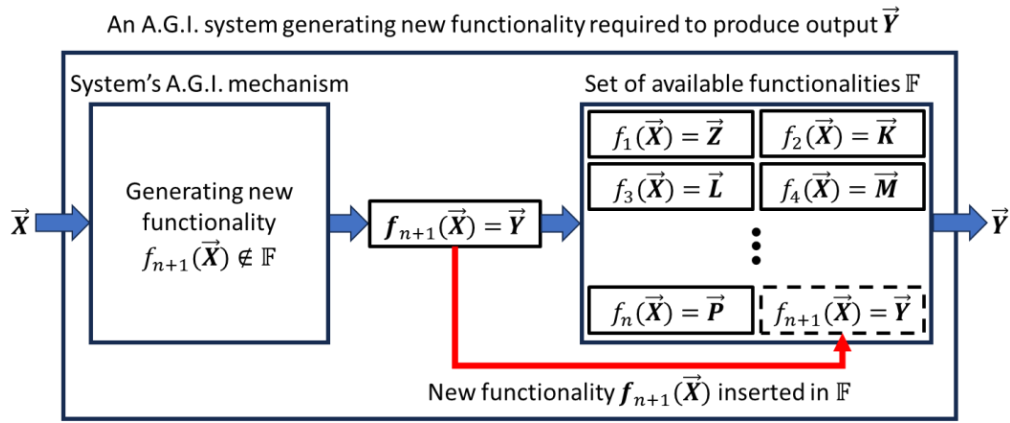

**Figure 1. A schematic representation of an A.G.I. system. An A.G.I. system has a mechanism that enables it to generate a new functionality $f_{n+1}(\vec{X}) = \vec{Y}$ that did not initially exist in the system's available functionalities, and is required to output some new knowledge $\vec{Y}$**

incentive is further magnified by the fact that, if such an algorithm is really feasible, then it is only a matter of time for it to be invented, and the first who does so, would be able to profit the most.

On the other hand, if it is impossible to create an A.G.I. algorithm, then heavily investing in its implementation could be wasteful both in research effort and capital. Instead, if we knew that A.G.I. is impossible to be achieved, we could readjust our vision of A.I. from a general purpose, autonomous A.G.I. model, to a narrower approach where a system is composed of a compilation of smaller models, each proficient on a single specific task (e.g., autonomous driving, image generation, etc.).

How we evaluate the dangers of A.I. would also change based on the two scenarios. For example, a highly autonomous A.G.I. model would be able to drive innovation based on its own priorities, that could potentially be misaligned with the priorities of the humans. Such fears, of an all-powerful A.I. that dictates the future of humanity have been expressed many times in the past [34, 35, 36]. Where, if no A.G.I. algorithm exists such concerns are no longer relevant. Thus, having a formal proof for or against the possibility of A.G.I. can be pivotal in our future decisions of how we approach this technology.

## 2. Background on Computability

In Section 1.1 we defined A.G.I. as the ability to create new functionality. Now we seek to answer the question, is it possible for an algorithm to achieve A.G.I. and thus, able to create new functionality that was previously not coded in the original algorithm? We can formalize this question by rephrasing it as follows, **"is the process of creating new functionality computable?"**. In this section we give a short background on computability theory and explain what it means for a process to be computable. In Section 3 we will use the concepts we present here to construct a formal proof and provide an answer to the above question.

### 2.1 Computability and the Church-Turing Thesis

A. Church [37] and A. Turing [38] pose and discuss the fundamental question of computability, i.e. which mathematical problems can be solved by a step-by-step (mechanical) process. Such a process, traditionally a function that maps the natural numbers on the natural numbers $f: \mathbb{N} \rightarrow \mathbb{N}$, is computable if there is an algorithm that can produce $f(n)$ for any given $n \in \mathbb{N}$. The function $f$ is required to be finite, i.e. it must finish in a finite number of steps, definite, i.e. each step must be precisely specified, deterministic, i.e. each next action must be determined by the current state, and it must always terminate, i.e. for any valid input the procedure must eventually halt.

Today the work of Church and Turing is widely described as the Church-Turing Thesis [39], which is a hypothesis about the nature of computation. It states that if a function can be calculated by an effective method, then it can also be calculated by a Turing machine [40], an abstract machine that at any given moment, its behavior can be completely determined by its current state and inputs. In other words, for every computable process (a.k.a. algorithm) there is a Turing Machine that can compute it. If there is no such machine, then the process is not computable. No algorithm can exist.

Furthermore, A. Turing introduced the Universal Turing Machine (UTM) [40], a machine that can compute any other Turing Machine. In other words, a UTM can compute all computable processes. Modern computers, are shown to be equivalent to the UTM and can be programmed to run different algorithms including emulating any other computing machine [40]. Thus, according to the Church-Turing Thesis, modern computers are able to compute all computable processes (given enough memory and time), a concept known as *Turing completeness.* Even though the Church-Turing thesis remains a conjecture, it still forms the basis of modern computer science.

### 2.2 Boolean Logic and the Universal Gates

Modern computers use Boolean logic to implement basic functions. The fundamental block of these machines is the Boolean gate. A 2-input Boolean gate implements a Boolean function $G_2(a, b) = r$, that takes as input two binary values $a$ and $b$, performs a logical operation, and outputs a single binary result $r$. Thus, there is a total of four input combinations (0,0), (0,1), (1,0), (1,1), where for each input the output can take a value of 0 or 1. This produces a total of 16 unique Boolean functions that are represented by 16 unique Boolean gates that perform different Boolean operations (AND, OR, NAND, NOR, etc.). A 1-input Boolean gate can be created simply by setting one of the inputs of the 2-input Boolean gate to a constant value, for example $G_1(a) = G_2(a, 1)$. Combining two of the 2-input Boolean gates can create a 3-input Boolean gate $G_3(a, b, c) = G_3(G_2(a, b), c)$. Continuing this process, we can generate Boolean gates with an arbitrary number of inputs.

Since a Boolean gate of an arbitrary number of inputs can be created, it follows that any Boolean function can be implemented using a combination of the 2-input Boolean gates. Boolean functions can then be used to implement all aspects of modern computers from arithmetic and logical units (ALUs), to networks, memory registers, control circuits, even the clock of the central processing unit (CPU) that synchronizes the operations of the whole system [41, 42]. Furthermore, it has been shown that the 2-input NAND gate (as well as the 2-input NOR gate) can be used to implement all other 2-input Boolean gates (the set of functions of the 2-input Boolean gates is exhaustive, i.e. there are exactly sixteen 2-input Boolean gates) [43]. That means that using only NAND gates, we can implement all Boolean functions and consequently all the components of a modern computer. That is why the NAND gate (as well as the NOR gate) is also referred to as a ***universal gate***. This is not just a theoretical concept but rather one can try this using online tools and

**Axiom 1:** For a process to be computable there must be a finite, integer, positive number ($\infty > k \geq 0$) of NAND gates that can be configured in a way that implements that process. If no such number $k$ exists, then the process is incomputable.

courses, where using just the NAND gate as a building block, they can design higher complexity circuits up to a fully functional general-purpose computer system [44, 45]. In practice, modern computer hardware may use a variety of gates and technologies to optimize design goals like latency and power. Yet, it is important to note here, that the notions described above describe a theoretical concept, i.e. the logical representation of Boolean functions, and they are not necessarily bound by the underlying technology that implements these functions.

From the above, as well as the Church-Turing Thesis we presented in Section 2.1 it follows that for every algorithm (i.e., computable process) there must be at least one configuration of a finite number of NAND gates that implements that algorithm. In other words, we can conclude that if no such configuration of finite number of NAND gates can be found for a given process, then that process can be deemed as incomputable and thus, no algorithm that implements that process can ever exist. We formalize this observation in **Axiom 1** and in Section 3 we use it as our starting point to reason about the computability of A.G.I.

## 3. Formal Proof on the Computability of A.G.I.

According to **Axiom 1** that emerges through the Church-Turing thesis as presented in Section 2, for a process to be computable, we must be able to map it to a logic circuit that uses a finite, integer number of NAND gates $k \geq 0$. If we can show that such a number $k$ exists, then a process is computable. If we can show that there is no such number $k$ then the process is not computable. We base our proof on this fundamental observation.

### 3.1 Proof by Contradiction

To investigate the computability of A.G.I. we start by first considering the simplest logical circuit we can imagine and then progressively build complexity. Thus, we can start with a single wire that simply carries an input to an output without affecting the value of its signal. Such a circuit would simply implement the function $f_w(a) = a$ as seen in **Figure 2**. We call this, the *single wire* function. The *single wire* alone does not implement an A.G.I. algorithm as defined in **Definition 1**, since it can never create *new functionality*. That is, the only functionality of the *single wire* is to pass the input to the output, unchanged. No other functionality can ever emerge. Although not a surprising conclusion we use this observation as a building block while we investigate if there is a circuit

A singe wire

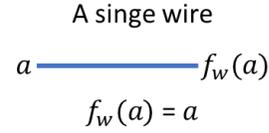

$f_w(a) = a$

| $f_w(a)$ | $a$ |
|---|---|
| 0 | 0 |
| 1 | 1 |

**Figure 2.** Using zero NAND gates ($k = 0$) to implement the *single wire* function $f_w(a) = a$

The NAND gate

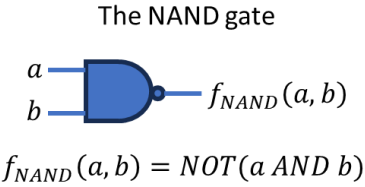

$f_{NAND}(a, b) = NOT(a \; AND \; b)$

| $f_{NAND}(a,b)$ | $a$ | $b$ |
|---|---|---|
| 1 | 0 | 0 |
| 1 | 1 | 0 |
| 1 | 0 | 1 |
| 0 | 1 | 1 |

**Figure 3.** The *NAND gate* function $f_{NAND}(a, b) = NOT(a \; AND \; b)$

that can implement an A.G.I. algorithm. We summarize this conclusion in **Lemma 1**.

The next step is to consider the universal NAND gate. As discussed in Section 2.2, it is sufficient to build any algorithm, no matter how complex, by using solely NAND gates. The NAND gate has a well-defined function $f_{NAND}(a, b) = NOT(a \; AND \; b)$ as seen in **Figure 3**. We refer to this function simply as the *NAND gate* function. A single, standalone NAND gate is not sufficient to implement an A.G.I. algorithm, since as shown in **Figure 3**, the *NAND gate* has a constant functionality, always returning the logical NAND operation of its inputs. Thus, no *new functionality* can ever be produced from a single NAND gate. We summarize this conclusion in **Lemma 2**.

As a next step in increasing complexity, we combine multiple *single wire* functions and one *NAND gate* function to create a circuit that takes $n$ input bits and returns $m$ output bits as seen in **Figure 4**. We call this the ***single NAND circuit*** denoted as $f_1(\vec{X}) = \vec{Y}$, where the subscript number on $f_1$ indicates the total number of NAND gates used to create the circuit, $\vec{X}$ is the $n$-bit input vector, and $\vec{Y}$ is the $m$-bit output

**Lemma 1:** The *single wire* function is not A.G.I. as it does not meet the requirements of Definition 1 for any input.

**Lemma 2:** The *NAND gate* function is not A.G.I. as it does not meet the requirements of Definition 1 for any input.

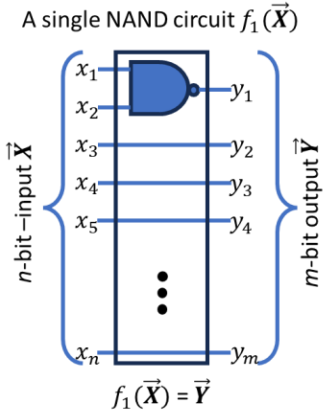

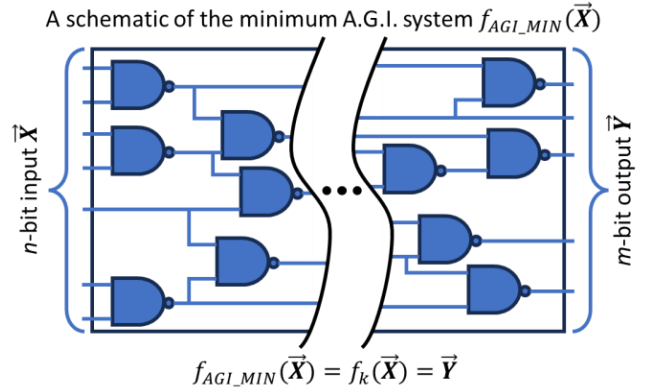

**Figure 4. A circuit that uses only a single NAND gate, has an *n*-bit input $\vec{X}$ and an *m*-bit output $\vec{Y}$, and denoted as $f_1(\vec{X}) = \vec{Y}$**

**Figure 5. An abstract schematic representation of a *k*-NAND circuit that implements the minimum (in terms of NAND gates) AGI algorithm $f_k(\vec{X}) = f_{AGI\_MIN}(\vec{X}) = \vec{Y}$.**

vector. Note that although **Figure 4** indicates that $m = n - 1$, this circuit can be used to implement any positive integer numbers of $n$ and $m$, by simply choosing how to use/connect the input and output wires of the circuit to actual input and output signals.

Carefully observing the *single NAND circuit* we can see that all its outputs can be expressed from the basic blocks we have already defined, the *single wire* and *NAND gate* functions. For example, the first output bit ($y_1$) from the output vector $\vec{Y}$ can simply be expressed as a function of the *NAND gate* $y_1 = f_{NAND}(x_1, x_2) = NOT(x_1 \text{ AND } x_2)$, where $x_1$ and $x_2$ are input bits of vector $\vec{X}$. The rest of the outputs bits from $\vec{Y}$ can be expressed from the *single wire* function $y_i = f_w(x_j) = x_j$ where $i > 1$ and $j = i + 1$. Thus we can conclude that the *single NAND* circuit will never produce *new functionality* but rather its output $\vec{Y}$ will always follow the functionality of the *single wire* and the *NAND gate* to produce results that we can exhaustively enumerate. Thus, once again, the *single NAND circuit* does not implement an A.G.I. algorithm. We summarize this conclusion in **Lemma 3**.

From the above we can conclude that if an A.G.I. algorithm exists then it must emerge as complexity increases. Such claims have been made before [46, 47, 48, 49], suggesting that A.G.I., similar to other computable capabilities, may emerge from higher complexity. If that is true, then from **Axiom 1** it follows that there will be a finite number of NAND gates, that can be used to create an A.G.I. algorithm. Moreover, there must also be a minimum number of NAND gates that are needed to implement the smallest (in terms of NAND gates used) A.G.I. algorithm. This is true because as we empirically know from simple algorithms that simply print a constant message (e.g., printing "Hello World" is not an A.G.I. algorithm) and as shown by **Lemma 1**,

**Lemma 2**, and **Lemma 3**, not all algorithms can achieve A.G.I. Let us refer to that algorithm as the ***minimum A.G.I system*** that can be implemented using $k > 0$ NAND gates in some configuration that achieves A.G.I. Thus, any algorithm that can be implemented using $l < k$ NAND gates cannot be A.G.I. We summarize this conclusion in **Lemma 4**.

Let us now assume that this ***minimum A.G.I. system*** exists and thus according to **Axiom 1** and **Lemma 4** it can be mapped in circuit that uses $k$ NAND gates. That means that we have a $k$-NAND gate circuit, denoted as $f_k(\vec{X}) = \vec{Y}$ that implements the ***minimum A.G.I system***, denoted as $f_{AGI\_MIN}(\vec{X}) = f_k(\vec{X}) = \vec{Y}$. As we do not know the value $k$ or the configuration of the NAND gates that implements the ***minimum A.G.I system*** we simply represent this circuit through a simplified, abstract schematic as shown in **Figure 5**. In **Figure 5** we have $k$ NAND gates that are connected through some unspecified configuration. The circuit gets as input an $n$-bit vector $\vec{X}$ and outputs an $m$-bit vector $\vec{Y}$. Observe, that every output bit from the vector $\vec{Y}$ *must either be an output of a NAND gate or a simple wire that directly carries an input bit to output.* This is because the wires, and the NAND gates are the only building blocks used. Thus, we can conclude that at least some of the output bits of $\vec{Y}$, are produced from the outputs of some NAND gates. This is true because the only other alternative is that all output bits come as direct wires from input bits and thus the circuit is actually using zero NAND gates and thus could not be implementing an A.G.I. algorithm.

Now, we can select one of the output bits that comes from a NAND gate (e.g., the top right NAND in the schematic shown in **Figure 5**) and "carve-out" that NAND gate in order to logically split our circuit into two distinct circuits as seen in

**Lemma 3: The *single NAND circuit* function is not A.G.I. as it does not meet the requirements of Definition 1 for any input.**

**Lemma 4: If an A.G.I. system exists, then a *minimum A.G.I. system* also exists. The *minimum A.G.I. system* is the system that uses the minimum amount $k > 0$ of NAND gates and still meets the requirements of Definition 1, to achieve A.G.I. Thus, any system that can be implemented using $l < k$ NAND gates cannot be A.G.I.**

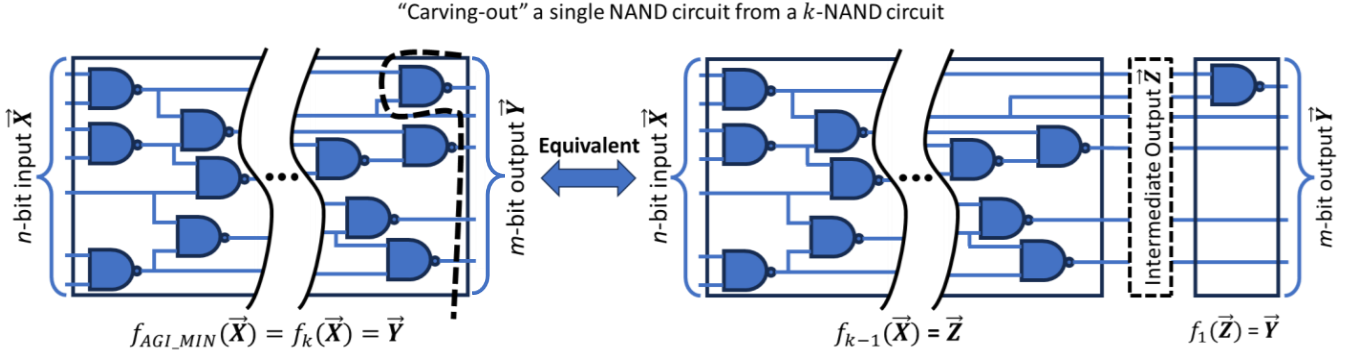

**Figure 6.** A schematic representation of how any *k*-NAND circuit $f_k(\vec{X}) = \vec{Y}$, can be expressed as a function of *k-1*-NAND circuit $f_{k-1}(\vec{X}) = \vec{Z}$ and a single-NAND circuit $f_1(\vec{Z}) = \vec{Y}$ so that $f_k(\vec{X}) = f_1(f_{k-1}(\vec{X})) = \vec{Y}$

**Figure 6**. Such a logical segmentation of a circuit does not impact the overall functionality of the original circuit and is a common practice that allows engineers to simplify the design process by focusing on smaller, simpler circuits that together assemble a larger more complex circuit [50, 51, 52]. After "carving-out" the new circuit, we have a $k-1$ NAND gate circuit on the left denoted as $f_{k-1}(\vec{X})$ that takes as input $\vec{X}$ and produces an intermediate output $\vec{Z}$. On the right we have a *single NAND gate* circuit, denoted as $f_1(\vec{Z})$ that takes as input the intermediate output $\vec{Z}$ and produces the final output $\vec{Y}$. Thus we can now conclude that our A.G.I. system can be also expressed as $f_{AGI\_MIN}(\vec{X}) = f_k(\vec{X}) = f_1(f_{k-1}(\vec{X})) = \vec{Y}$. Note, that to come to this conclusion we made no assumptions for the size of the input ($n$), or output ($m$), or how the $k$ NAND gates are configured and connected with each other.

Now, if $f_{AGI\_MIN}(\vec{X}) = \vec{Y}$ represents an A.G.I. system, then according to **Definition 1**, there must be some input $\vec{X}$ for which the A.G.I. system produces some ***new functionality***. However, $f_{AGI\_MIN}(\vec{X}) = f_1(f_{k-1}(\vec{X})) = \vec{Y}$ for every input $\vec{X}$. This means that $f_1(f_{k-1}(\vec{X})) = \vec{Y}$ should also be an A.G.I. system. Thus the *new functionality* was either produced from $f_{k-1}(\vec{X}) = \vec{Z}$ (a circuit that uses $k-1$ NAND gates) or $f_1(\vec{Z}) = \vec{Y}$ (a circuit that uses a single NAND gate). However, from **Lemma 4** any system that uses less than $k$ NAND gates cannot be A.G.I. Thus, both these functions cannot produce ***new functionality*** no matter their input. This means that the entire system $f_1(f_{k-1}(\vec{X})) = \vec{Y}$ also fails to create ***new functionality*** and thus fails to meet the requirements of **Definition 1**. This leads to a logical contradiction and therefore we can conclude that:

***There is no finite, positive number k of NAND gates that can be used to implement an A.G.I. algorithm, as defined in Definition 1. Thus, as follows from Axiom 1, A.G.I. is incomputable!***

## 4. Discussion

In this work we provided a formal proof that A.G.I., as defined by many pioneers of the industry [21, 22, 23, 27, 28, 29, 30, 31, 32] is an incomputable process. In this section we discuss the implications of this proof as it relates to future A.I. development, how we understand the dangers of A.I., as well as what that means for the origin of human intelligence. Furthermore, we discuss the limitations of our proof and how they could be addressed in the future.

### 4.1 Implications

#### a. Impact on A.I. Developement

Given that A.G.I. is an incomputable process, we should consider how that may impact the development of A.I. technologies. We start by emphasizing that A.I., as we previously discussed in Section 2, has already showed many promising technological advancements in areas like autonomous driving, medical diagnosis, image generation, and many other disciplines [2, 3, 4, 5, 6, 7, 8, 9]. These advancements, were feasible in the absence of A.G.I. Moreover, one can argue that an A.I. model without A.G.I. may be more suitable for some implementation (e.g., autonomous driving) where we want a model to follow strict rules imposed by the model's designers, rather than having its own agency that may lead to safety concerns. Thus even without A.G.I., A.I. as a technology is still very impactful.

However, we can conclude that efforts to create a fully autonomous, general, A.G.I. model, that would be able to self-advance science and technology, cannot be materialized. Thus we should reconsider such efforts and examine alternative options in A.I. development. One such option, could be to focus on more narrow A.I. models that specialize in smaller more specific tasks. This is an approach that can help us design more accurate A.I. models, that can better assist humans in narrower fields rather than a general purpose approach.

#### b. Reevaluating the Dangers of A.I.

The proof we present in this work, that A.G.I. is incomputable, also reveals that fears of an all-powerful A.I. model that could develop its own functionalities, make its own decisions and defy the functional restrictions of its design [34, 35, 36] are no longer substantiated. However, that does not mean that no other dangers can emerge from the A.I. technology.

For instance, overstating and overestimating the capabilities of A.I. can lead to misconceptions and misuse. Misinterpreting an A.I. model as A.G.I., may lead users to assume that they can automatically trust A.I. as a highly reliable, or even infallible source of truth. For example, cases of people being misguided to assume that A.I. models are sentient and thus can help them answer the "life's mysteries" have been recently reported [53].

Similar can be the case in the scientific community, where trusting A.I. as a truly intelligent source can lead to ***information stagnation and a slowdown in scientific advancement***. As A.I. cannot generate new functionality or new information, we must realize that in the best case, A.I. can only provide us with the current knowledge frontier (i.e., the current state of human knowledge). As with every point in history, the current knowledge frontier also contains inaccurate or incomplete understandings that only a truly G.I. system like humans can further advance. Thus, it is of the utmost importance to understand that the human creativity and innovation cannot be replaced by A.I. models or by any other type of algorithm.

### c. Origins of Human's General Intelligence

Another important implication that arises from our proof is that human's G.I., like A.G.I., must be an incomputable process (as they both describe the similar capabilities). That raises the following questions, how are humans able to achieve G.I.? What is the origin of human G.I.?

To answer these questions, one must first consider two possibilities, that all physical processes can also be computed by a Turing machine or maybe there are some physical processes that are incomputable. The Physical Church-Turing thesis (Physical CT) [54] links mathematical notions to the physical world, associating physical (mechanical, digital or quantum) computing devices with the theoretical concept of a Turing machine. G. Piccinini [55] expands on the computability of physical processes and argues that not all physical processes can be considered physical systems. Instead, he claims that the Physical CT can be divided into two categories, the "Bold Physical CT" and a more relaxed "Modest Physical CT". The Bold Physical CT argues that all physical processes can be computed by Turing machines, where the Modest Physical CT argues that only physical processes that represent compute systems (a.k.a. physical functions) can be computed by Turing machines. With the advent of digital computers and the rapid advancements of information theory since the original work of Church and Turing, a common agreement (although yet unproven) has settled among researchers in favor of the Modest Physical CT.

Based on this understanding on physical process computability, there are only two possible outcomes. The first possibility is that human's G.I. could be a non-computing physical process and thus incomputable by a Turing machine. The only other possibility is that human's G.I. is a non-physical process. This conclusion could potentially open a new frontier for exploration on human intelligence that could help us better understand how human's G.I. works.

Another implication is that human's G.I., as an incomputable process, it cannot itself be a result of a computable process. That means, that no algorithmic process can be the source of the human's G.I. This conclusion may come in conflict with our modern understanding of human evolution. Human evolution is understood and described as step-by-step algorithmic process [56, 57, 58] that iteratively, through generations allowed humans to develop their current features and traits. However, at least one of these traits, human's G.I., cannot be a product of an algorithmic procedure. This observation could lead us to reevaluate our understanding of the evolution process and how it relates to the origin of human intelligence.

### 4.2 Limitations

In this work we presented a formal proof that A.G.I. is an incomputable process. Our proof is based on the Church-Turing thesis which is considered the cornerstone of modern computer science and computability theory. However, the Church-Turing thesis (although widely accepted as true) is itself not proven. Thus, if the thesis was ever disproven, our proof will also have to be re-examined. In this section, we explore and discuss that possibility.

As we discussed in Section 2.1, the Church-Turing thesis postulates that all "effectively computable" methods can be computed by a Turing Machine. However, the definition of what constitutes an "effectively computable" method is not well-defined. Moreover, we currently do not know how many such methods exist in total. Thus, it is very difficult (if not impossible) to prove the claim of this thesis. If an "effectively computable" method was ever discovered that cannot be executed by the Turing Machine then the Church-Turing thesis would be refuted.

Thus, another possible interpretation of this work, is that A.G.I. as we define it in **Definition 1** is indeed an "effective computable" method but not all such methods can be computed by Turing Machines. Such an interpretation would essentially claim to refute the Church-Turing thesis. That would mean that at least theoretically there must be some Hyper-Turing Machine (HTM), with novel computation mechanisms, that can compute such methods that a Turing Machine cannot. To achieve that, a HTM would perform computations that go beyond our current understanding and imagination. Note that no Turing Machine (and by extension any modern or future computer that implements a Turing Machine) would ever be able to emulate the behavior of an HTM. Moreover, even if theoretically such a HTM exists, that does not necessarily mean that it is possible to physically implement it.

We emphasize though, that in this work we do not claim to refute the Church-Turing thesis but rather that A.G.I. is an incomputable process. Regardless of how our proof is interpreted, the basic conclusion will remain true, that no Turing Machine can compute A.G.I., no matter the underlying technology used to implement it.

## 5. Related Work

Although, to the best of our knowledge, no other work presents a formal proof on the computability of A.G.I., prior work has attempted to investigate the subject as it relates to computability and the reasoning of A.I. models. One such work comes from A. Keller [59] who uses Gödel's incompleteness theorem [60] to argue that A.I. models, like all other algorithms and complex systems, are mathematically incomplete. As the author explains, systems can only produce a finite set of results and can be used to prove a finite set of postulates. Indeed, every system uses axioms, that itself cannot prove. Thus, no system can be truly general as one would expect from an A.G.I. model. The author further strengthens his argument by giving an analogy with Kolmogorov complexity [61]. Kolmogorov complexity proves that no general algorithm that provides optimal solution for all problems can ever be conceived. One such example are compression algorithms, where depending on the nature of the input data, different compression algorithms achieve different compression rates. The best choice of a compression algorithm simply depends on the input's bit pattern. No general compression algorithm that achieves optimal compression for all inputs can ever exist. Thus, the author argues, no truly general A.G.I. model can ever be achieved.

To build his argument, the author assumes that an A.G.I. model must be able to solve every conceivable problem optimally. However, this not possible even for humans that are indeed G.I. Moreover, the total human knowledge that can ever be achieved may also be finite, and thus, even humans may not meet the "truly general" requirement that the author sets for A.G.I. As a matter of fact, the author concludes that his work may not necessarily prove that A.G.I. is impossible but rather that it indicates that A.I. can never be more intelligent that humans.

Similar arguments are made by M. M. Schlereth [62] that also uses Gödel's incompleteness theorem [60] and Kolmogorov complexity [61] to argue that no truly general A.G.I. algorithm is possible. He also uses Rice's theorem [63], to argue that many aspects of complex systems are known to be incomputable, and thus A.G.I. may also be incomputable. Rice's theorem states that non-trivial properties of Turing Machines, for example asking questions about their semantic properties, are incomputable. One famous such example is the Halting Problem were asking if an algorithm (or Turing Machine) ever halts is an undecidable problem as proved by Alan Turing [38]. However, no precise explanation is given why that should imply that A.G.I. must be incomputable. The author adds one more argument against A.G.I. computability by claiming that algorithmic systems cannot increase uncertainty and thus increase entropy as defined by Shannon [64]. Instead, with each new input the uncertainty of the system only decreases and thus entropy decreases as well. The author states that a truly "thinking" system, when receiving some new information, also raises new questions that should lead to entropy to increase once again. The author refers to that phenomenon as entropy divergence. However, the claim that entropy always decreases in algorithmic systems is not entirely true. For example, consider a decompression algorithm that takes an input message and returns a decompressed, larger in bit-length output message. Such an algorithm effectively increases entropy. In contrast a compression algorithm decreases entropy. However, in both cases, the meaningful information that is contained in the message (the functionality and knowledge that the message conveys) remains the same regardless if the message is compressed or decompressed. No new knowledge is ever created. Shannon's entropy, when describing information, does not refer to the actual meaning (e.g., knowledge or functionality) contained in a message but rather the maximum theoretical information that a message could potentially transmit, even if in practice a message has no real meaning (e.g., it contains only gibberish).

Other prior work has tried to use a more empirical approach by investigating how specific A.I. models work and thus analyze if they could potentially achieve A.G.I. A work from Apple researchers [65] analyzed how reasoning models "think" and evaluated their results for different problems that have different levels of complexity. The authors demonstrate that, although reasoning models can excel in medium complexity problems, they perform no better than smaller non-reasoning models for lower complexity problems. Furthermore, for higher complexity problems both reasoning and non-reasoning models fail. The authors thus conclude that the capabilities of reasoning models may be overstated and thus may not be path to A.G.I. Research from Anthropic [67] has also focused on the reasoning steps of LLMs. Their research has shown that the reasoning steps are often inconsistent in respect to the produced result. For example, the reasoning may be correct, but the result is unrelated to the reasoning steps and thus wrong, and vice versa. This indicates that LLMs' reasoning is not close to what one would expect for an A.G.I. model. Another work [66], analyzed the results of LLMs and showed that these models are unable to generalize beyond their training data. Framing this within the context of our work, LLMs cannot produce functionality and knowledge that was not included in their training data. Finally, P. V. Coveney et al. [68] argue that LLMs face a "scaling wall". More specifically, they assert that as LLMs attempt to scale to increase generality their accuracy decreases.

Although these observations on LLMs and their reasoning steps are interesting (and they align with what we would expect based on our proof on A.G.I. non-computability) they only pertain to specific A.I. models. Thus, one can always argue that maybe a different model type, size, or other algorithm may be able to solve these issues and eventually achieve A.G.I. In contrast, in this work, we prove that not just todays A.I. models, but rather no A.I. model or algorithm can ever achieve A.G.I. At least not without a ground breaking revision on the fundamentals of computer science and computability theory.